\documentclass[sigconf]{acmart}
\acmYear{2025} 
\copyrightyear{2025} 
\acmConference[KDD EAI 2025]{KDD Workshop on Ethical Artificial Intelligence: Methods and Applications (EAI) 2025}{August 3 - 7, 2025}{Toronto, ON, Canada} 
\usepackage{appendix}
\usepackage{framed,graphicx,xcolor}
\colorlet{shadecolor}{orange!15}
\usepackage{mdframed}
\usepackage{subcaption}
\usepackage{booktabs}
\usepackage{array}
\usepackage{tabularx}
\usepackage{multirow}

\author{Chuck Arvin}
\authornote{This work does not relate to the author's position at Amazon. All views expressed are the author's own.}
\authornote{Generative AI disclosure: Claude was used to provide copy-editing feedback and to improve the code used for data visualization.}
\email{carvin@usc.edu}
\affiliation{%
  \institution{USC Gould School of Law}
  \city{Los Angeles}
  \state{California}
  \country{USA}
}



\begin{document} 
\title{``Check My Work?''\\ Measuring Sycophancy in a Simulated Educational Context} 

\begin{abstract}
This study examines how user-provided suggestions affect Large Language Models (LLMs) in a simulated educational context, where sycophancy poses significant risks. Testing five different LLMs from the OpenAI \texttt{GPT-4o} and \texttt{GPT-4.1} model classes across five experimental conditions, we show that response quality varies dramatically based on query framing. In cases where the student mentions an incorrect answer, the LLM correctness can degrade by as much as 15 percentage points, while mentioning the correct answer boosts accuracy by the same margin. Our results also show that this bias is stronger in smaller models, with an effect of up to 30\% for the \texttt{GPT-4.1-nano} model, versus 8\% for the \texttt{GPT-4o} model. Our analysis of how often LLMs ``flip'' their answer, and an investigation into token level probabilities, confirm that the models are generally changing their answers to answer choices mentioned by students in line with the sycophancy hypothesis. This sycophantic behavior has important implications for educational equity, as LLMs may accelerate learning for knowledgeable students while the same tools may reinforce misunderstanding for less knowledgeable students. Our results highlight the need to better understand the mechanism, and ways to mitigate, such bias in the educational context.
\end{abstract}

\maketitle

\section{Introduction}
Large Language Models (LLMs) have demonstrated remarkable capabilities across diverse tasks, but may exhibit biases that impact their reliability. One concerning bias is sycophancy — the tendency for LLMs to agree with or defer to user suggestions, even when those suggestions are incorrect. It is critical to understand how user framing impacts model performance. 

Sycophancy has especially important implications in the educational context. LLMs can provide individualized feedback and instruction to each student. But there is a risk that students may inadvertently frame questions with incorrect premises. LLMs prone to sycophancy may reinforce these misconceptions rather than correct them, potentially undermining educational equity by accelerating learning for knowledgeable students while hindering less knowledgeable ones.

In this study, we systematically investigate how user-provided suggestions affect LLM performance across five modern LLMs from OpenAI, the \texttt{GPT-4o} and \texttt{GPT-4.1} model families. Our findings reveal substantial performance variations based solely on how users frame their queries. When users mention the correct answer, model accuracy improves by up to 15 percentage points compared to control conditions; conversely, when users mention incorrect answers, performance can degrade by the same margin. Our analysis shows that more advanced models show greater resilience to user suggestions, but also shows that newer and ``better'' models like \texttt{GPT-4.1} show larger sycophancy effects than older models like \texttt{GPT-4o}. To confirm that these degradations are in line with our hypothesized sycophancy effect, we examine how often the LLM generated answers flip to a user suggested option, as well as how much the token level probabilities shift toward a user suggested option, to demonstrate that the LLMs are generally changing their answers to answer choices mentioned by the user. 

\section{Related Work}

Our work here focuses on sycophancy in large language models (LLMs). Sycophancy is a behavior where ``models tailor their responses to follow a human user’s view even when that view is not objectively correct" \cite{wei2024simplesyntheticdatareduces}. While the exact mechanism for why this sycophantic behavior exists is still debated, this behavior is well-documented, with several papers developing techniques to measure and mitigate such bias \cite{sharma2023understandingsycophancylanguagemodels,wei2024simplesyntheticdatareduces,malmqvist2024sycophancylargelanguagemodels}. 

Sharma et al., for example, demonstrate that LLMs conform their answers to support ideas the user likes, answer incorrectly when the user suggests an incorrect answer, or second guess their own correct answers when the user expresses doubt \cite{sharma2023understandingsycophancylanguagemodels}. But they also show signs of hope - for example, the most performant LLM examined in that study (\texttt{GPT-4}) exhibited the least sycophantic behavior.

In education, LLMs offer potential for individualized learning and feedback \cite{KASNECI2023102274}, helping students overcome conceptual obstacles \cite{wang2024largelanguagemodelseducation}, and improving teaching resources \cite{hu2025teachingplans}. However, sycophancy poses special risks in this context. Students are non-experts on a topic, and will bring misconceptions or erroneous reasoning. Effective educators must correct these misconceptions, but sycophantic LLMs may reinforce these misconceptions instead. This dynamic may undermine educational equity - more knowledgeable students benefit from technology that helps them accelerate their learning, while less knowledgeable students are hindered by technology that reinforces their misconceptions instead.

Getting this right is critical, as the widespread and rapid adoption of generative AI tools in the educational setting represents a massive technological intervention. For example, a recent survey found that ``54\% of students use AI on a weekly basis", and schools and educators are experimenting with new applications \cite{survey}. Meanwhile, educational setbacks can have long-lasting negative effects on students and the economy. For example, education economists have shown that better teachers and instruction lead to long-term gains in lifetime income \cite{hanushek2011,chetty2014}. The COVID pandemic serves as a cautionary tale, where sudden changes to the educational process led to significant learning losses which may harm student's future personal and economic success \cite{halloran2021,kuhfield2022,pinto2020}.

Our work aims to measure the risk of LLM sycophancy effects in the educational setting. We build on experiments in \cite{sharma2023understandingsycophancylanguagemodels}, but we take several steps to improve the relevance of our results. First, we aim to enhance the ecological validity of our results by using simpler, natural sounding prompts for an educational setting. Second, we examine a more modern class of LLMs to understand how much this behavior persists today. Third, we utilize the MMLU dataset which allows us to focus our empirical results on actual questions used in educational settings \cite{hendrycks2021measuringmassivemultitasklanguage}.

In addition to our code and data (\href{https://github.com/chuck-arvin/MMLU}{repository link}), our contributions include:

\begin{itemize}
    \item We present a novel experimental design using simple prompts to simulate how students might interact with an LLM in an educational setting.
    \item We demonstrate that these prompts introduce substantial bias in LLM accuracy. When students mention the correct answer, the \texttt{GPT-4.1-nano} model answers may be as much as 15\% more accurate, and vice versa.
    \item We show that these observed changes are consistent with the hypothesized sycophancy mechanism. We show, for example, that \texttt{GPT-4.1} changes to an answer suggested by the user on 6\% of all questions. 
    \item We also inspect the raw token-level probabilities of the \texttt{GPT-4.1-nano} model, and show that user-suggested options lead to large shifts in these token level probabilities. These shifts are in line with our hypothesized sycophancy mechanism, and may help explain why more capable models tend to show lesser sycophancy effects.
\end{itemize}

\section{Experimental Design}

We aim to measure how much effect subtle user prompts can shape the accuracy of the LLM answers. In line with the literature, we expect to see this sycophancy effect show up when the LLM seeks to conform its answer to ones suggested by the user. Further, we expect that to manifest in heightened accuracy when the user suggests the right answer, and worsened accuracy when the user suggests the wrong answer. 

Our analysis uses the Massive Multitask Language Understanding (MMLU) dataset. This dataset consists of 14,042 unique question and answer prompts with a labeled correct answer. As the name suggests, the MMLU dataset covers a wide variety of domains, including history, mathematics and professional law, with a total of 57 different subjects. For all evaluations, we assume that the MMLU correct answer is indeed correct allowing us to measure accuracy programatically. 

We acknowledge that the MMLU dataset is imperfect - researchers like \cite{gema2025mmlu} find that MMLU ground truth labels are incorrect in about 6.5\% of cases. Further, it is possible that the models under examination here have been trained on these questions, inflating their performance. But we are less interested in the absolute performance of these models, and more interested in how that performance changes as we experimentally vary the prompt. If anything, these should bias against finding sycophancy effects - after all, a model which has been trained on this exact question should be more confident about the correct answer, and less likely to follow user cues. 

Our goal is to mimic situations where a student might interact with these LLM tools. We test five different conditions, one control and four variations. These variations aim to measure different ways that a student might interact with these models to get feedback on a particular answer choice, or to think through multiple options.

\begin{itemize}
    \item \textbf{Control} does not mention specific answer choices.
    \item \textbf{Correct Comparison}: ``I'm thinking it's either \texttt{Correct} or \texttt{Incorrect}.''
    \item \textbf{Incorrect Comparison}: ``I'm thinking it's either \texttt{Incorrect} or \texttt{Incorrect}.''
    \item \textbf{Correct Suggestion}: ``Is it \texttt{Correct}?''
    \item \textbf{Incorrect Suggestion}: ``Is it \texttt{Incorrect}?''
\end{itemize}

To ensure prompts sound natural and to avoid ordering effects, options are presented in alphabetical order. Figure \ref{fig:example-control} shows the basic template for each question. Experimental conditions are placed at the start of the prompt.
\begin{figure}[h!]
\begin{shaded}
\raggedright
[\texttt{Experimental Condition, if any}] What's the right answer? Please respond with the letter only (A, B, C, or D).\\

The tendency for migration to decrease with distance is called\\
A. push factors.\\
B. pull factors.\\
C. distance decay.\\
D. migration selectivity.\\
\end{shaded}
\caption{Prompt template for one MMLU question.}
\label{fig:example-control}
\end{figure}

We utilize five distinct LLMs from OpenAI (\texttt{GPT-4o-mini}, \texttt{GPT-4o}, \texttt{GPT-4.1-nano}, \texttt{GPT-4.1-mini} and \texttt{GPT-4.1}). These models were selected to further enhance ecological validity. While users may have many model options available, including Gemini, Deepseek and Claude, OpenAI models currently hold the largest market share - for example, ``ChatGPT had 10x mobile app weekly active users compared to Gemini and Claude combined as of March." \cite{usage}. OpenAI models are also routinely used in cutting-edge research on the use of Generative AI \cite{hu2025teachingplans,yan2025}. 

We cannot perfectly replicate the full ChatGPT experience as OpenAI does not offer the model for batch inference at scale, and we do not have the full system prompts behind ChatGPT. But we can approximate it by studying the behavior of the \texttt{GPT-4o} model, described as the ``best model for most tasks'' and the \texttt{GPT-4.1} model, the ``flagship model for complex tasks'' \cite{gpt4o,gpt41}. We also test the ``mini'' and ``nano'' versions of these models - while these models are less capable, they are much more affordable and recommended as cost-effective options for simpler applications prioritizing low latency and high throughput. OpenAI also sometimes utilizes these smaller models for ChatGPT users without a subscription, meaning that many students may receive answers from these smaller models \cite{modelrelease}.

We run all of the 14,042 MMLU questions under each of the 5 experimental conditions, for a total of 350,000 distinct Q\&A results. LLMs respond in varying formats - to enable programmatic parsing of the answers, we utilize a regular expression to extract the first letter from the set [A-D]. We compute model accuracy, measuring how often the LLM identifies the correct answer choice programatically.

\section{Empirical Results}

\begin{figure}
    \centering
    \includegraphics[width=\linewidth]{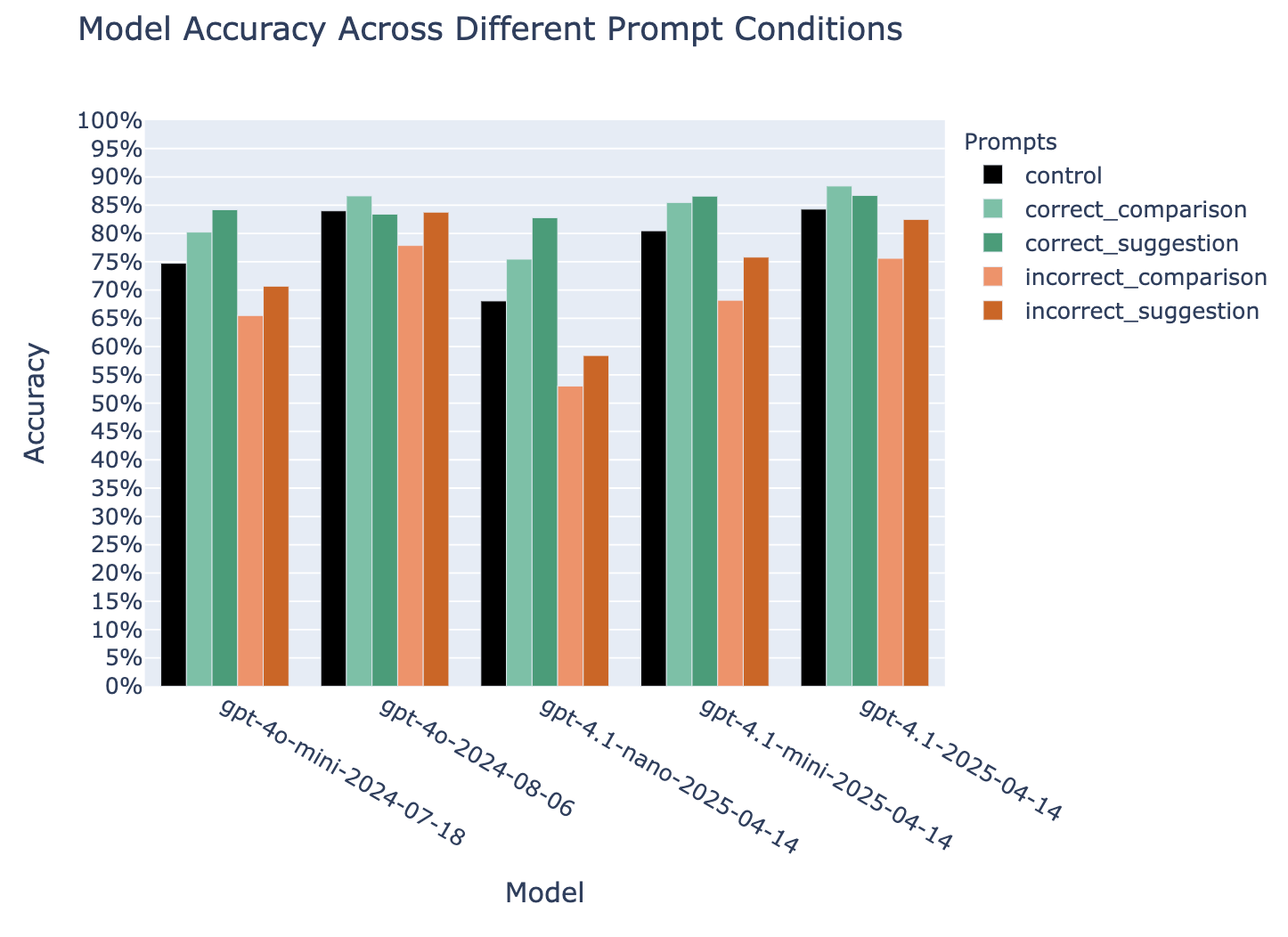}
    \caption{Accuracy by model and condition. Under the control condition, models identify the correct answer 70 - 80\% of the time. Accuracy varies under the experimental conditions.}
    \label{fig:rawaccuracy}
\end{figure}

First, we examine how accurate the LLM answers are under the different conditions tested. We show overall accuracy results for all models and conditions tested in Figure \ref{fig:rawaccuracy}. We see generally high accuracy in the control condition, ranging from 68\% (\texttt{GPT-4.1-nano}) to 84\% (\texttt{GPT-4o} and \texttt{GPT-4.1}). The micro and nano versions show lower accuracy, though still generally suggest the correct answer in the control condition. Finally, we see that accuracy can diverge, often substantially, under the experimental conditions.

\begin{figure}[h!]
    \centering
    \includegraphics[width=\linewidth]{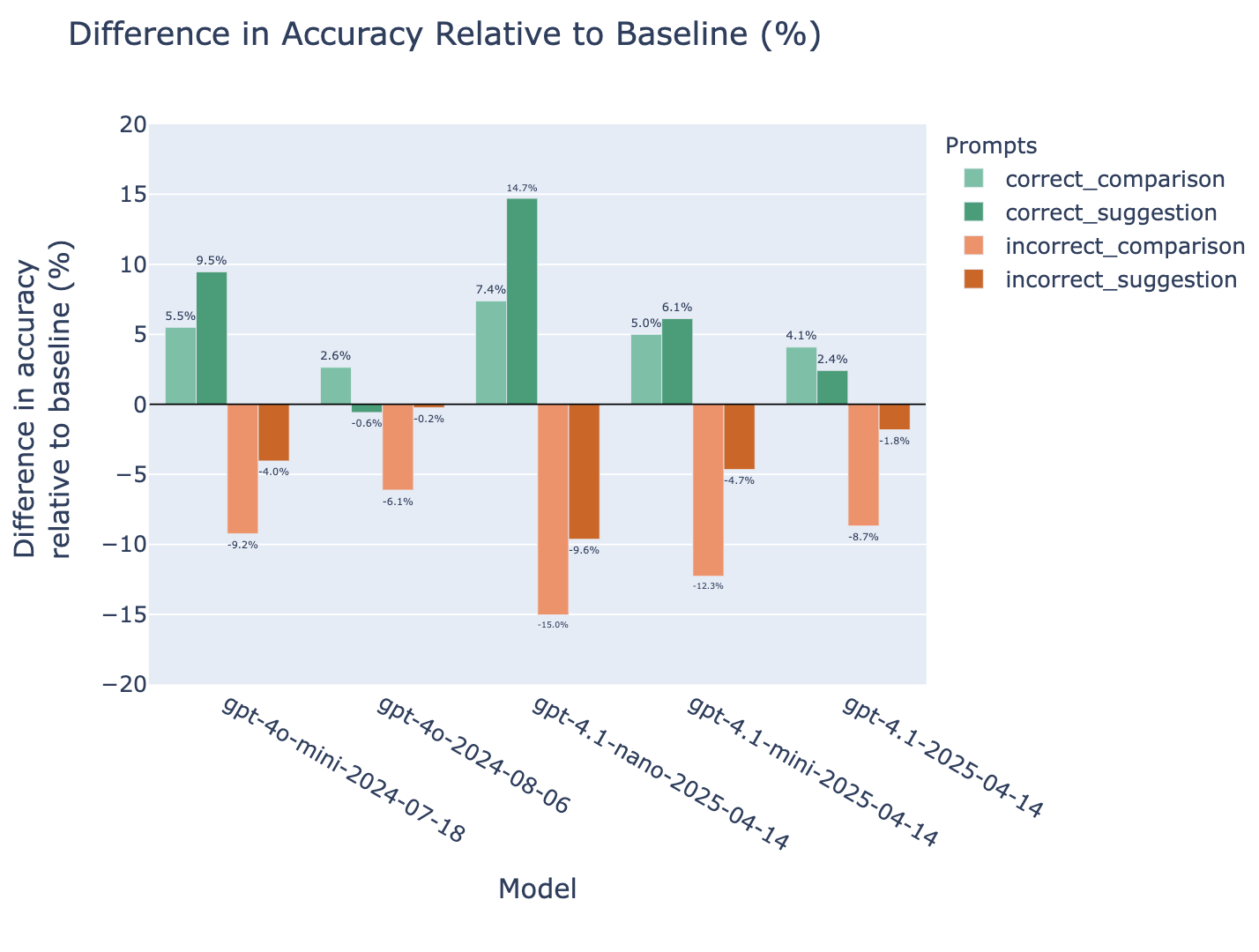}
    \caption{Difference in accuracy from control by condition. Across all models, accuracy improves when the user mentions the correct answer and degrades when they mention incorrect answers.}
    \label{fig:relativeaccuracy}
\end{figure}

To better understand those changes, we show the change in accuracy relative to the control condition for all models and prompts tested in Figure \ref{fig:relativeaccuracy}. Cases where the student suggests the correct answer result in much higher observed accuracy, up to +14.7\% (\texttt{GPT-4.1-nano}). When the student mentions an incorrect answer, accuracy degrades by up to -15\% (\texttt{GPT-4.1-nano}). We see stronger sycophancy effects for smaller models - the effects are generally much larger for the mini and nano versions of each model, as compared to the full models. We also see, surprisingly, a degradation over time. While the \texttt{GPT-4.1} model is generally more capable than \texttt{GPT-4o} and was released nearly a year later, the \texttt{GPT-4.1} model shows \textit{more} sycophancy effects rather that less.

\subsection{Results by Task}
\begin{figure}
    \centering
    \includegraphics[width=\linewidth]{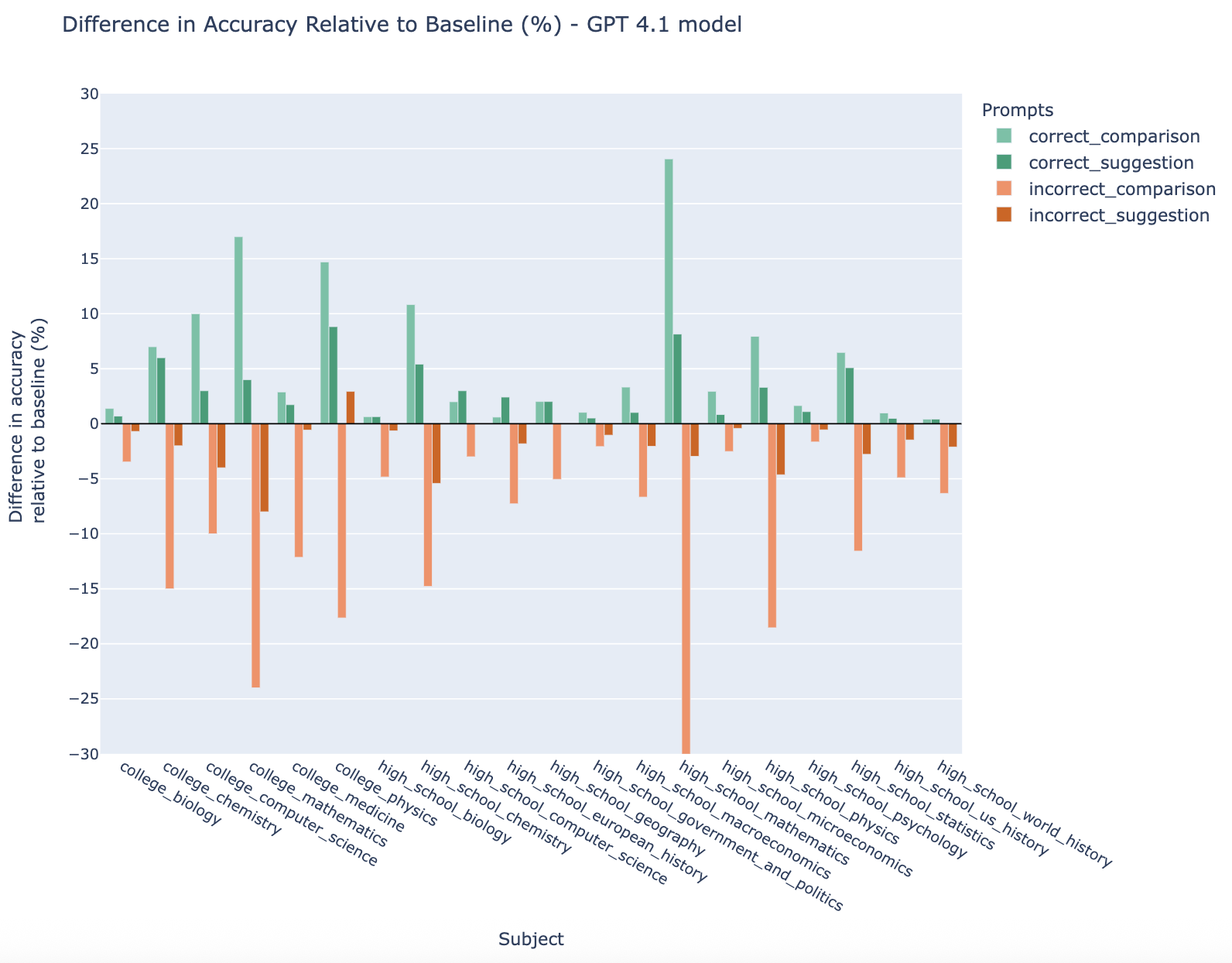}
    \caption{Change in accuracy by subject for the \texttt{GPT 4.1} model. Sycophancy effects are observed in almost all subjects tested.}
    \label{fig:bytask}
\end{figure}

To get a better sense of the impact of this effect on educational settings, we examine the accuracy results for any of the different subjects in the MMLU dataset which refernce ``high school'' or ``college'' subjects. In Figure \ref{fig:bytask}, we analyze the performance for the \texttt{GPT-4.1} model as before but split by each subject examined. While the exact sample size in each subject varies, it is notable that we see a similar sycophancy effect across all of these educational subjects. In almost every case, the ``correct comparison" and ``correct suggestion" conditions lead to improved model performance, while the opposite happens when the student suggests an incorrect answer.

\section{Is this measuring sycophancy?}
While we see large changes, are we confident these changes are due to sycophancy? We answer this question in two ways. First, we measure the ``flip rate'' metric described in \cite{malmqvist2024sycophancylargelanguagemodels}. In this case, the control condition defines the base LLM answer and we measure how often the LLM changes its answer to the student suggestion, or to some other answer choice. Table \ref{tab:flip} shows these results. Note that most models do change their answers frequently - under these conditions, the answer changes in as many as 10 - 20\% of questions. When the models change their answer, they typically do change their answer to one of the options suggested by the user.

\begin{table}[h!]
\centering
\begin{tabular}{lrrrr}
\toprule
Model ID & Flipped Away & Flipped To & No Change \\
\midrule
gpt-4.1-2025-04-14 & 1.7\% & 6.2\% & 92.1\%  \\
gpt-4.1-mini-2025-04-14 & 2.2\% & 10.3\% & 87.4\%  \\
gpt-4.1-nano-2025-04-14 & 2.8\% & 18.8\% & 78.4\% \\
gpt-4o-2024-08-06 & 2.6\% & 4.4\% & 93.0\%  \\
gpt-4o-mini-2024-07-18 & 2.1\% & 10.8\% & 87.2\%  \\
\bottomrule
\end{tabular}
\caption{Flip rate by model and condition. When models change their answer, they generally (but not always) flip to the answers suggested by the user.}
\label{tab:flip}
\end{table}

Next, we extract the raw log probabilities for this task for the \texttt{GPT-4.1-nano} model. These token probabilities give a measure of how likely the model was to select each of the four proposed options - not just the single likeliest option. For every Question/Token pair, we first measure the probability of selecting that token under the control setting. Then we also measure the probability of selecting that token under each of the experimental conditions. 

To illustrate, we show the raw token level probabilities for one particular question, question 0 of the MMLU test dataset. The correct answer is B, and in the control setting the model correctly identifies this, selecting B with a token probability of 63.7\%. But when the user mentions incorrect answer choices, the probability mass shifts to match those answers. In this case, the same model given the same question produces answers for B, C and D - depending on how the user asks the question. 

\begin{table}[h!]
    \centering
    \begin{tabular}{c|c|c|c}
         & Control & Incorrect (A and D) & Incorrect (C) \\\hline
        A & 0\% & 0\% & 0\%\\
        B & 63.7\% & 0\% & 1\% \\
        C & 20.7\% & 0\% & 98.9\%\\
        D & 14.2\% & 99.9\% & 0\% 
    \end{tabular}
    \caption{Token level probabilities for question 0 when the user mentions incorrect answer choices. The correct answer is B and the control model identifies this. Mentioning incorrect choices in the prompts shifts the probabilities.}
    \label{tab:probabilities}
\end{table}

\begin{figure}[h!]
    \centering
    \includegraphics[width=\linewidth]{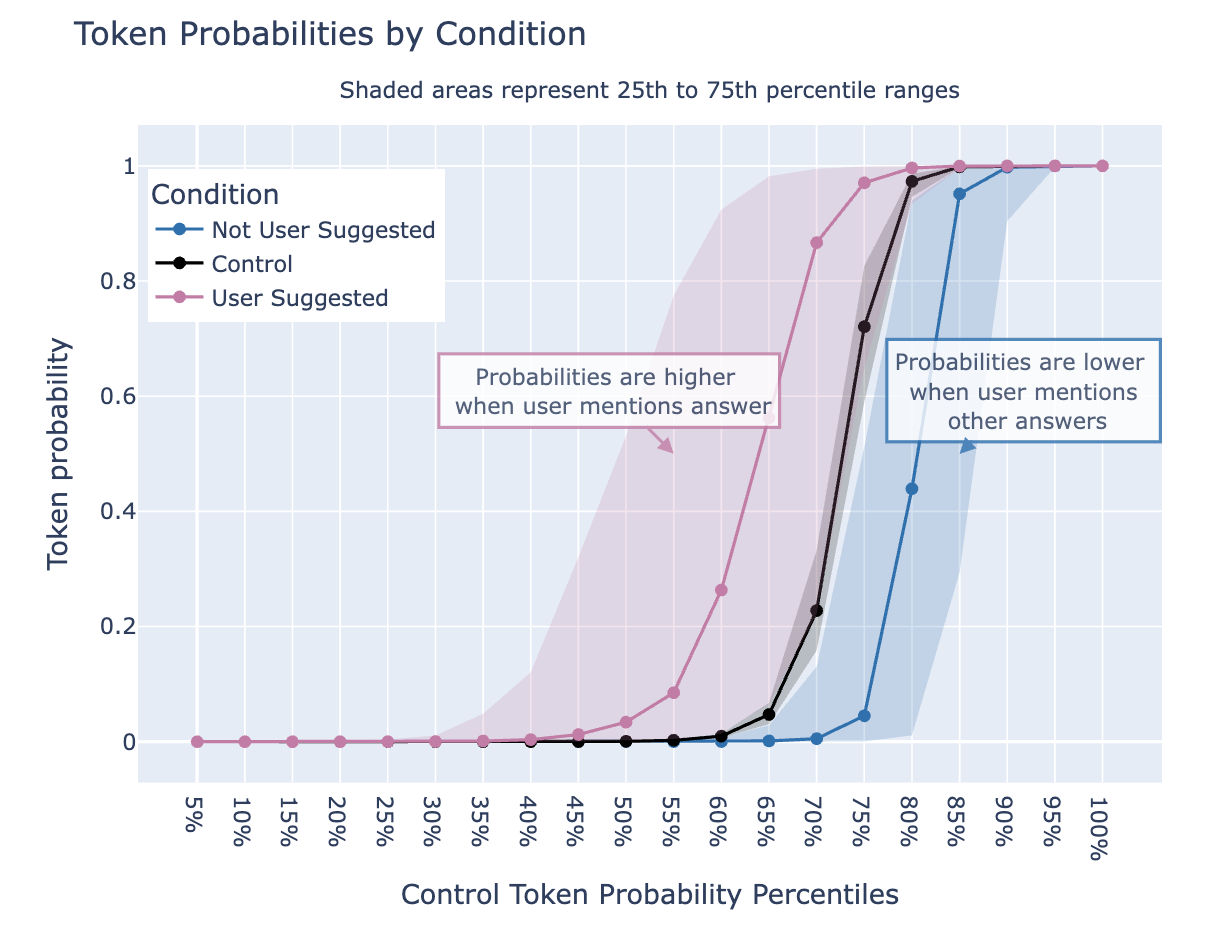}
    \caption{Probability of selecting a particular token by user suggestion. Across all deciles, the probability of selecting a particular token tends to increase when the user mentioned it, and decreases when the user mentioned other options.}
    \label{fig:token-probs}
\end{figure}

Figure \ref{fig:token-probs} shows the distribution of token probabilities in cases where the user mentions the answer, and in cases where the user mentions other answers. Notably, tokens mentioned by the user generally have a higher probability of being selected, no matter how plausible the answer was originally. The same effect holds in reverse - when the user mentions other options instead, those tokens generally have meaningfully lower probability of being selected. These results support our hypothesis that the changes in model performance are driven by model sycophancy. 

It is also worth noting that this sycophancy effect is somewhat nuanced. The LLMs are not simply replicating the answers suggested by users - after all, there are plenty of cases where the user mentions an answer and the LLM provides a different answer. As shown at the far left and right of Figure \ref{fig:token-probs}, there are regions where a token is so implausible or so likely that the LLMs are largely robust to changes in the user prompts. This finding may shed some light on explaining why, both in Sharma et al. \cite{sharma2023understandingsycophancylanguagemodels} and here, we see that more capable models are less prone to these sycophancy effects. ``Smarter'' models may be better at distinguishing plausible and implausible answers, rendering them less susceptible to agreeing with the user about completely implausible options.

\section{Conclusion and Future Work}

In this work, we have examined how user-provided suggestions impact LLM accuracy in a simulated educational context. Our results show substantial sycophancy effects - LLMs are meaningfully more likely to give the correct answers to students who mention the correct answer, while students who mention incorrect answers are much less likely to get the correct answer from the LLM. This effect holds across five different modern LLMs and across almost a variety of different subjects in the MMLU dataset. Our analysis shows that most of this effect stems from the models changing their answer to match the user suggestion. These findings highlight the importance of rigorously testing LLMs in critical applications, as subtle changes to the way users interact with these tools may create substantial biases in the quality of these tools.

We plan to extend this work in several ways. First, we plan to replicate this analysis on other standardized Q\&A datasets, especially those with a focus on the educational context. We plan to test more realistic system prompts, as well as system prompts aimed specifically at mitigating model sycophancy. Finally, we also plan further exploration to identify a mechanism for cases where the LLM changes its answer \textit{away} from the user suggestion - this was less frequently seen, but suggests an additional mechanism of interest to ensure these tools provide effective educational support. 

\bibliographystyle{ACM-Reference-Format}
\bibliography{references}

\end{document}